\documentclass[]{infiniai}
\usepackage{microtype}
\usepackage{graphicx}
\usepackage{booktabs}
\usepackage{array}
\usepackage{tabularx}
\usepackage{amsmath}
\usepackage{subcaption}
\usepackage{textcomp}
\usepackage{wrapfig}
\usepackage{xurl}          
\usepackage{fontawesome5}     
\DeclareRobustCommand{\linkicon}[2]{\makebox[1.5em][l]{#1}#2}  

\usepackage{caption}
\colorlet{RefFill}{infinibg}
\colorlet{WinnerFill}{infinitint}
\makeatletter
\newcommand{\keepspace}[1]{\par\begingroup\dimen@=\dimexpr\pagegoal-\pagetotal\relax\ifdim\dimen@<#1\newpage\fi\endgroup}
\makeatother

\usepackage[capitalize,noabbrev]{cleveref}

\title{ServerBench: How can your coding agents design specialized LLM serving systems?}

\author{Xinrui~Zhong$^{1,*}$, Zhiming~Wang$^{1,*}$, Yongqi~Liu$^1$, Zhaohui~Wang$^2$, Shanglin~Yang$^1$,\\ Zikun~Li$^1$, Zhuoming~Chen$^1$, Beidi~Chen$^1$}
\contact{\{xinruizh, zhimingw, yongqil\}@andrew.cmu.edu, zwang000@usc.edu, \\ \{shangliy, zikunl, zhuominc, beidic\}@andrew.cmu.edu}
\affiliation{$^1$Carnegie Mellon University\quad $^2$University of Southern California}
\contribution[*]{Equal contribution.}
\metadata[\linkicon{\faGithub}{Code}]{\href{https://github.com/Infini-AI-Lab/Server-Bench}{\texttt{github.com/Infini-AI-Lab/Server-Bench}}}
\metadata[\linkicon{\faBlog}{Blog}]{\href{https://infini-ai-lab.github.io/Server-Bench/}{\texttt{infini-ai-lab.github.io/Server-Bench}}}

\abstract{Large language model (LLM) coding agents develop software through
experimentation, offering an opportunity to design the serving systems
supporting their own operation. Effective serving requires coordinating
implementation and operating decisions under session dependencies,
latency constraints, and correctness requirements. We introduce
ServerBench, a benchmark for agentic design of complete, specialized
LLM serving systems. Agents experiment against a common objective,
\emph{dollars per dollar}: latency-qualified token value per dollar of GPU cost. Stateful replay preserves session
dependencies, and independent verification checks the submitted system.
Across 11 coding-agent models and three deployments, Claude Sonnet
5.5 achieves 2.97 dollars per dollar on Gemma E4B, $4.87\times$ the vLLM
reference; 22 of 33 runs yield verified systems,
and 13 exceed the strongest baseline. Our analysis identifies three findings: high-value designs
coordinate capacity improvements with load selection; agents interpret
performance feedback through workload-relevant experiments; and reliable delivery requires
verification of the execution paths introduced by optimization.
Overall, ServerBench supports systematic study of how agents translate service
objectives and feedback into effective, verified serving
systems for given task specifications.
}

\hypersetup{
  pdftitle={ServerBench: How can your coding agents design specialized LLM serving systems?},
  pdfauthor={Xinrui Zhong; Zhiming Wang; Yongqi Liu; Zhaohui Wang; Shanglin Yang; Zikun Li; Zhuoming Chen; Beidi Chen}
}

\begin{document}

\maketitle

\section{Introduction}
\label{sec:intro}
\begin{figure}[h]
\centering
\includegraphics[width=\textwidth]{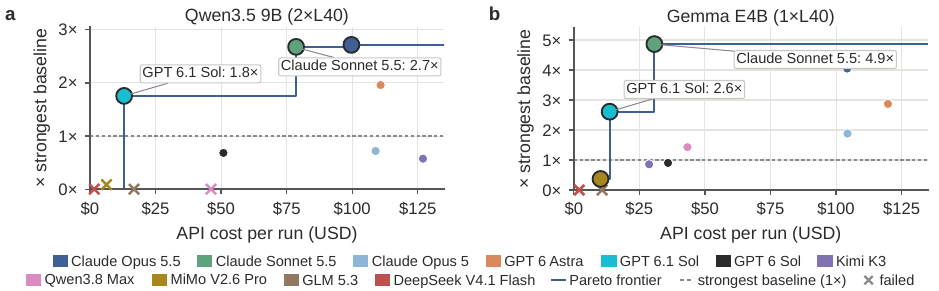}
\caption{API cost and serving value relative to the strongest baseline.
GPT 6.1 Sol offers lower-cost gains; Claude reaches higher value.
Costs use OpenRouter rates (\Cref{tab:api-costs}).}
\label{fig:cost-value}
\end{figure}

Large language model (LLM) coding agents build on code-generation
capabilities \citep{zheng2025livecodebenchpro,zhou2026autocode}, using tools
and execution feedback to undertake software development and professional
workflows \citep{yang2024swe,xiao2025csr,chan2025mle,sun2026ale}. These
workflows depend on LLM serving systems, which run models and handle
requests \citep{kwon2023efficient,zheng2024sglang,zhou2026janus}. This creates an
opportunity for coding agents to design these systems. Service providers
need correct model execution and timely responses. We therefore ask:
\emph{how can coding agents design LLM serving systems while
satisfying given deployment and service constraints?}

\begin{figure*}[t]
\centering
\includegraphics[width=\textwidth]{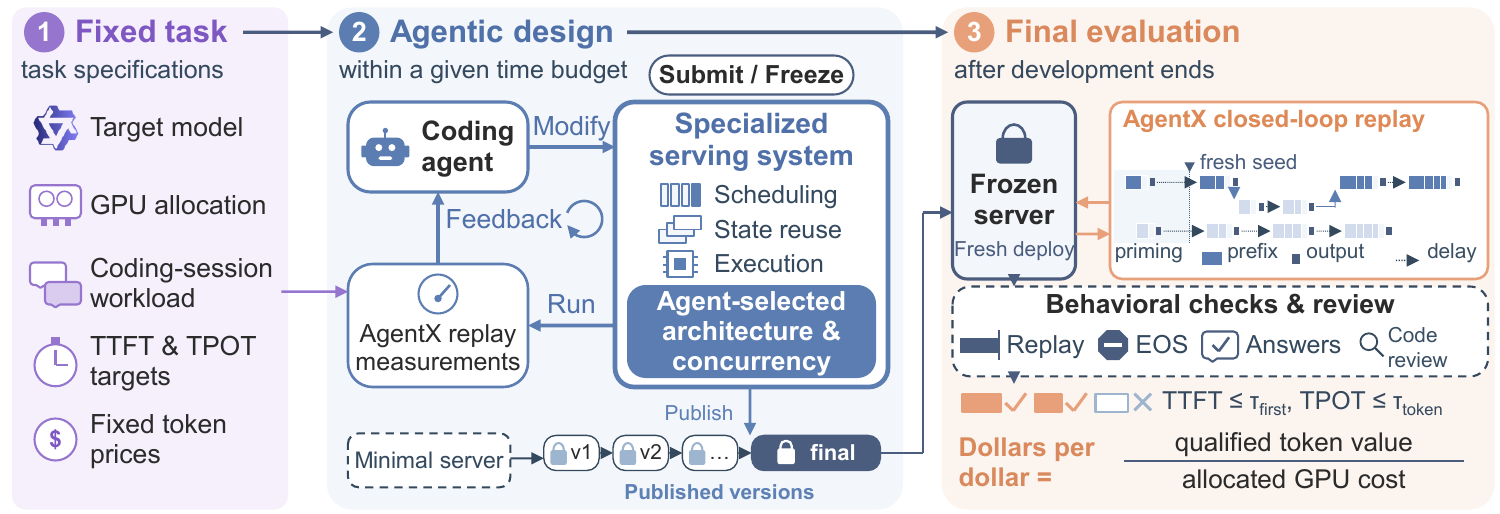}
\caption{Agents design the server and choose how many sessions it serves
at once for a given model, GPU allocation, and workload. Replay preserves
shared context and dependencies between requests; final checks assess
model correctness and complete workload~execution.}
\label{fig:overview}
\end{figure*}

Despite rapid progress in agent-driven systems optimization, existing
benchmarks evaluate individual kernels, code patches, or servers under
preset test conditions
\citep{ouyang2025kernelbench,nangia2026iso,williams2026swe,yeon2026inferencebench,xu2026seis},
leaving three gaps in evaluating how agents design complete serving systems:
\begin{itemize}
\item \textbf{Oversimplified workloads.} Prior benchmarks issue
independent requests or fixed prefill, decoding, and throughput
scenarios \citep{yeon2026inferencebench,xu2026seis}. With uniform request lengths
and fixed batches, a static batch serves nearly as well as a production
engine, so techniques that real deployments depend on, such as
continuous batching and paged attention
\citep{yu2022orca,kwon2023efficient}, bring little benefit. Real
applications such as coding agents instead mix short and long requests,
extend a shared context over many turns, and fan out into parallel
subagents, where prefix reuse, memory management, and prefill--decode
interference shape performance
\citep{zheng2024sglang,srivatsa2025preble,agrawal2024taming,zhong2024distserve}.
\item \textbf{Speedup-only objectives.} Existing benchmarks score speedup
or throughput relative to a baseline
\citep{ouyang2025kernelbench,yeon2026inferencebench,xu2026seis}. Such objectives
ignore latency requirements and GPU cost and fix the offered load,
whereas a service provider must decide how much workload to serve
while keeping time to first token (TTFT) and time per output token
(TPOT) within targets.
\item \textbf{Correctness-only validation.} Validation typically checks
outputs against a reference implementation or test suite
\citep{ouyang2025kernelbench,williams2026swe,xu2026seis}. A serving system must
additionally meet tail-latency targets on the deployed workload and
execute that workload in full, with complete prompts, prescribed output
lengths, and consistent streaming. Optimizations such as speculative
decoding, state reuse, and custom scheduling introduce execution paths
that short or isolated tests rarely exercise.
\end{itemize}

We introduce \textbf{ServerBench}, a benchmark in which coding agents
build a complete LLM serving system for a given model, GPU allocation,
and workload, and decide how much workload it serves
(\Cref{fig:overview}). Starting from a minimal Transformers server,
agents plan experiments and revise their systems within a fixed
development budget. ServerBench closes the three gaps with three designs.
\textbf{1.~Workload: real coding-agent trajectories.} ServerBench
replays recorded coding-agent sessions \citep{semianalysis2026agentx}
with their request lengths, multi-turn structure, shared prefixes,
parallel subagent branches, and inter-turn delays. Requests wait on the
responses they depend on, so serving latency governs how much work
completes within the measurement window (\Cref{sec:workload}).
\textbf{2.~Objective: dollars per dollar.} We price the input and
output tokens of each completed request at fixed API rates, count only
requests that meet both TTFT and TPOT targets, and divide by the GPU
cost of the measurement window (\Cref{sec:score}). For example, USD 3.65
of qualified token value over USD 0.83 of GPU cost scores 4.40. The
objective rewards capacity improvements together with a matching load
choice: serving more workload raises gross value only while requests
still meet their latency targets.
\textbf{3.~Validation: beyond correctness.} Latency targets are the
p90 TTFT and TPOT of an SGLang reference with its model-specific
configuration, frozen before development (\Cref{sec:calibration}). The
submitted system is rebuilt in a fresh deployment and replayed on
held-out seeds. Independent checks verify complete prompts, output
lengths, streaming consistency, generation fidelity to the checkpoint,
and held-out answer accuracy; a post-run review inspects code and
recorded actions for prohibited shortcuts (\Cref{sec:validity}). Only
verified systems receive a score.

We evaluate eleven coding-agent models across three model--hardware
deployments. Claude Sonnet 5.5 achieves 2.97 dollars per dollar on Gemma
E4B, $4.87\times$ the vLLM serving baseline
(\Cref{sec:main-results}). Trajectory analysis identifies three
findings.
\emph{High-value designs coordinate capacity and concurrency.}
Sonnet 5.5 improves its server before raising concurrency, then limits
load when additional sessions worsen TPOT
(\Cref{sec:loop-analysis}). \emph{Interpreting feedback in workload
context guides optimization.} A random-text probe leads one agent to reject an
optimization; workload replies reveal its benefit and lead another to
adopt it
(\Cref{sec:reward-analysis}). \emph{Verification must match the
deployed workload.} Agents' short or isolated checks can miss long-context
stalls and cross-request crashes; known faults can persist or recur after
edits in submitted systems
(\Cref{sec:verification-analysis}).

Our contributions are:
\begin{itemize}
\item ServerBench, a benchmark for designing complete LLM serving
systems to maximize service value per GPU dollar
(\Cref{sec:task}). Agents design the server and set the number of
concurrent application sessions.
\item An evaluation protocol that replays coding sessions with shared
context and dependent requests, supports experiments designed by agents,
and independently verifies the final system
(\Crefrange{sec:workload}{sec:validity}).
\item A study of eleven coding-agent models across three deployments,
using controlled comparisons and trajectory analysis to explain
successes and failures in system design, experimental judgment, and
verification (\Crefrange{sec:main-results}{sec:ablation}).
\end{itemize}

\keepspace{6\baselineskip}   
\section{Related Work}
\label{sec:related-work}


\subsection{Inference System Optimization}
Orca and vLLM improve request scheduling and memory management
\citep{yu2022orca,kwon2023efficient}. SGLang and Preble exploit prefix reuse
\citep{zheng2024sglang,srivatsa2025preble}, while Sarathi-Serve and DistServe
mitigate prefill--decode interference through chunked prefill and prefill--decode
disaggregation \citep{agrawal2024taming,zhong2024distserve}.
AlphaEvolve and AI-Driven Research for Systems (ADRS) use evaluator
feedback to generate and refine algorithms and system code
\citep{novikov2025alphaevolve,cheng2025barbarians}.
Vortex enables agents to generate and refine sparse-attention algorithms
in a programmable serving system \citep{chen2026vortex}.
ServerBench studies whether agents integrate such optimizations into systems under fixed
deployment conditions.

\subsection{Inference-Performance Benchmarks}
MLPerf Inference standardizes workloads, quality targets, and deployment scenarios
\citep{reddi2020mlperf}. Agent-oriented benchmarks assess correctness and
performance: KernelBench tests generated GPU kernels against PyTorch references
for correctness and speedup \citep{ouyang2025kernelbench}, ISO-Bench compares
vLLM and SGLang optimization patches with expert implementations
\citep{nangia2026iso}, and $\Phi$-Bench spans kernel completion through
end-to-end infrastructure optimization \citep{ding2026phi}. SWE-Serve tests
production features within SGLang \citep{williams2026swe}, while InferenceBench
evaluates complete servers on prefill, decoding, concurrent throughput,
and a balanced scenario using baseline-relative speedup
\citep{yeon2026inferencebench}. ServerBench extends this line of evaluation
to response-dependent, stateful sessions and a latency-qualified
revenue-to-cost objective, linking implementation and operating choices
to the service delivered by complete systems.

\section{ServerBench}
\label{sec:serverbench}
ServerBench evaluates agentic design through a service objective,
stateful workload replay, iterative experimentation, and final verification.
Task configurations appear in \Cref{app:calibration,app:agent-configuration}.

\subsection{Task and Service Objective}
\label{sec:task}
Each task fixes a model checkpoint, GPU allocation, coding-session
workload, and service requirements. Starting from a minimal Transformers
server $S_0$, an agent has a time budget $B$ to develop a serving system.
Its submission comprises an implementation $S_B$ and session concurrency
$c_B$. Agents may redesign execution, batching, memory management, scheduling,
and state reuse \citep{yu2022orca,kwon2023efficient,zheng2024sglang}
under the prescribed checkpoint, precision, and inference semantics.
Agents prioritize changes to meet the service objective.

\paragraph{Dollars per dollar.}
\label{sec:score}
The objective, \emph{dollars per dollar}, measures completed token value
meeting both latency targets per dollar of GPU cost. For request $i$,
client-measured input and output counts $x_i,y_i$ give
$v_i=(p_{\mathrm{in}}x_i+p_{\mathrm{out}}y_i)/10^6$, with prices in USD
per million tokens. Let $q_i=1$ when its finite TTFT and TPOT meet targets
$\tau_{\mathrm{first}},\tau_{\mathrm{token}}$, and zero otherwise.
A verified system measured for $W$ seconds after warmup on $d$ GPUs,
priced at $r$ dollars per GPU-hour, receives
\begin{equation}
 D(S_B,c_B)=\frac{\sum_{i\in\mathcal{R}}q_i v_i}{d\,r\,(W/3600)},
 \label{eq:objective}
\end{equation}
where $\mathcal{R}$ contains completed requests. Define gross value
$G=\sum_{i\in\mathcal{R}}v_i/[d\,r\,(W/3600)]$ and the value-weighted
qualified share $f=D/G$ for $G>0$; thus $D=Gf$.
Input value includes cached prefixes at full price once per logical request. For example, USD 3.65 of qualified
value over USD 0.83 of GPU cost yields 4.40.

\paragraph{Latency calibration.}
\label{sec:calibration}
Before development, we select a reference operating point using SGLang's model-specific cookbook configuration
\citep{zheng2024sglang}. At the reference operating point, repeated
replays yield p90 TTFT $a_j$ and TPOT $b_j$ for seed $j$.
We set $\tau_{\mathrm{first}}=\max_j a_j$ and
$\tau_{\mathrm{token}}=\max_j b_j$, then freeze these request-level
targets before agent development. Taking the maxima provides headroom
for observed run-to-run variation. Agents choose the concurrency used by their submitted server.

\subsection{Stateful Workload and Closed-Loop Replay}
\label{sec:workload}
Coding sessions contain dependent turns, shared context, and parallel
subagent branches. Using the AgentX harness \citep{semianalysis2026agentx},
we replay response-dependent turns and parallel branches using recorded
inter-turn delays, with long global idle periods compressed.
Offered concurrency bounds active top-level sessions, whose branches
can issue additional requests. Response times thus affect session
progress and the work completed within the measurement window.

Privacy-preserving reconstruction replaces prompt and tool contents
while retaining request lengths, shared prefixes, and session topology.
Tasks fix reconstructed prompts and output lengths. Shared context
supports computation reuse \citep{zheng2024sglang,srivatsa2025preble},
while retained state and active requests compete for memory. Agents are
responsible for translating these workload characteristics into optimizations
while preserving model behavior and executing the workload in full.

Growing context enables reuse, while new sessions add prefill work
that competes with decoding \citep{agrawal2024taming,zhong2024distserve}.
Session progress thus combines prompt processing, generation, and retained state.

\FloatBarrier
\subsection{Agent Environment and Development Loop}
\label{sec:development}
\begin{wrapfigure}[18]{r}{0.45\linewidth}
\centering
\includegraphics[width=0.88\linewidth]{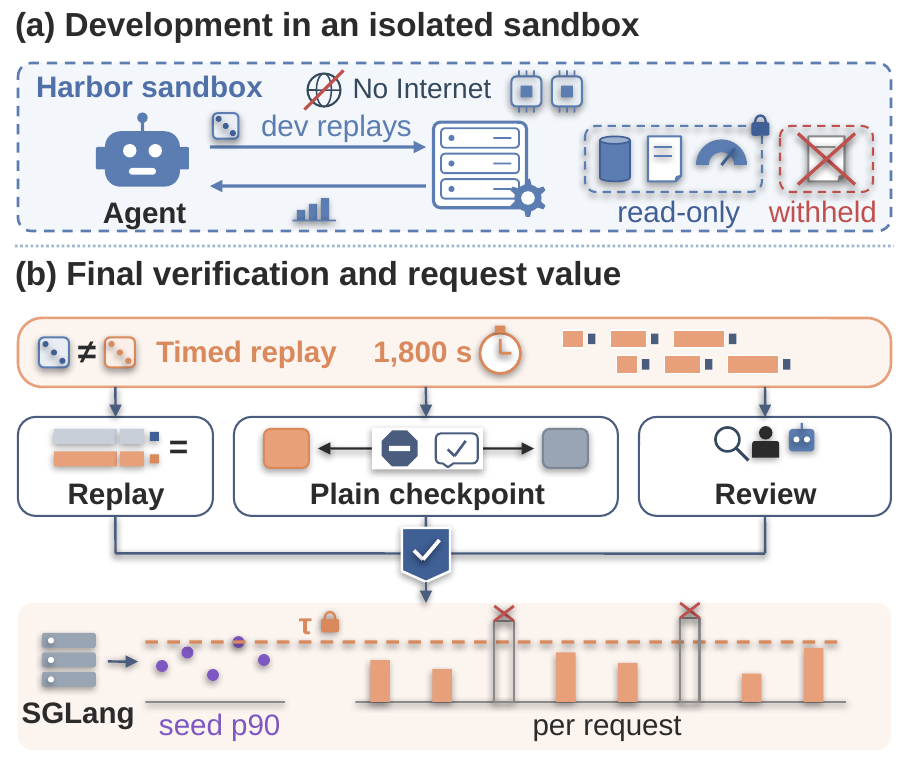}
\caption{Stateful replay connects session progress to the agent's design
experiments (a). Final evaluation verifies the published system and measures
request value against calibrated latency targets (b).}
\label{fig:interface}
\end{wrapfigure}
Agents work in an Internet-disabled Harbor sandbox with container or
Modal backends \citep{merrill2026terminal}. They receive a minimal server,
read-only model weights, reference configurations, measurement tools, and
the published performance procedure. Task instructions require model fidelity
and complete workload execution and prohibit evaluation manipulation.
Independent correctness-test details are withheld, and trusted evaluation
components remain outside the editable serving implementation.

The tools report latency, throughput, and request value. Agents organize
their own experiments, diagnose bottlenecks, and revise the implementation
\citep{yang2024swe}. They budget time for replays, diagnostics, and development.
Publishing fixes the implementation and concurrency separately from
performance testing; when the allotted time for development expires,
the last published version becomes the submission.

Agents choose experiment scale: diagnostics isolate suspected bottlenecks,
while complete replays test interacting requests. They decide which
observations warrant further tests, which changes to retain, and when to
publish. Versioned submissions connect these decisions to the implementation
evaluated after~development.

\begin{table*}[!t]
\centering
\footnotesize
\colorlet{RefFill}{infiniline!45}
\colorlet{WinnerFill}{infinitint}
\setlength{\tabcolsep}{2.5pt}
\renewcommand{\arraystretch}{1.20}
\begin{tabular*}{\textwidth}{@{\extracolsep{\fill}}l*{3}{rrr}@{}}
\toprule
\textbf{System} & \multicolumn{3}{c}{\textbf{Gemma E4B} ($1\times$L40)} & \multicolumn{3}{c}{\textbf{Qwen3.5 9B} ($2\times$L40)} & \multicolumn{3}{c}{\textbf{Qwen3.8 27B INT4} ($2\times$L40)}\\
\cmidrule(lr){2-4}\cmidrule(lr){5-7}\cmidrule(lr){8-10}
\rowcolor{RefFill}
\textbf{Best baseline} & \multicolumn{3}{c}{\textbf{vLLM} \$/\$ = 0.61, $c$ = 8} & \multicolumn{3}{c}{\textbf{SGLang} \$/\$ = 1.62, $c$ = 1} & \multicolumn{3}{c}{\textbf{vLLM} \$/\$ = 1.47, $c$ = 1}\\
Other baseline & \multicolumn{3}{c}{SGLang \$/\$ = 0.16, $c$ = 2} & \multicolumn{3}{c}{vLLM \$/\$ = 1.49, $c$ = 1} & \multicolumn{3}{c}{SGLang \$/\$ = 1.42, $c$ = 4}\\
\midrule
 & \textbf{\$/\$ @ $c$} & \textbf{TTFT} & \textbf{TPOT} & \textbf{\$/\$ @ $c$} & \textbf{TTFT} & \textbf{TPOT} & \textbf{\$/\$ @ $c$} & \textbf{TTFT} & \textbf{TPOT}\\
\cmidrule(lr){2-4}\cmidrule(lr){5-7}\cmidrule(lr){8-10}
Claude Opus 5.5 & 2.47\,@40 & 3.01 & 118.59 & \cellcolor{WinnerFill}\textbf{4.40\,@16} & 2.25 & 28.16 & \cellcolor{WinnerFill}\textbf{2.64\,@8} & 11.44 & 46.40\\
Claude Sonnet 5.5 & \cellcolor{WinnerFill}\textbf{2.97\,@32} & 2.75 & 83.05 & 4.34\,@20 & 12.34 & 26.76 & 2.01\,@12 & 46.57 & 47.35\\
GPT 6 Astra & 1.75\,@32 & 4.77 & 94.19 & 3.18\,@12 & 15.62 & 21.94 & 1.48\,@4 & 12.96 & 42.83\\
GPT 6.1 Sol & 1.59\,@24 & 4.43 & 72.45 & 2.84\,@4 & 2.71 & 23.67 & 0.98\,@4 & 75.66 & 43.31\\
Claude Opus 5 & 1.15\,@24 & 3.98 & 109.12 & 1.16\,@1 & 5.51 & 28.09 & 0.62\,@6 & 33.55 & 62.97\\
GPT 6 Sol & 0.55\,@5 & 3.63 & 110.52 & 1.11\,@1 & 13.19 & 20.45 & 0.26\,@2 & 136.48 & 49.53\\
Kimi K3 & 0.52\,@6 & 8.67 & 55.32 & 0.93\,@1 & 4.99 & 33.12 & \multicolumn{3}{c}{\textit{Not passed}}\\
Qwen3.8-Max-0902 & 0.87\,@14 & 2.61 & 59.16 & \multicolumn{3}{c}{\textit{Not passed}} & \multicolumn{3}{c}{\textit{Not passed}}\\
MiMo V2.6 Pro & 0.22\,@10 & 43.63 & 108.96 & \multicolumn{3}{c}{\textit{Not passed}} & \multicolumn{3}{c}{\textit{Not passed}}\\
GLM 5.3 & \multicolumn{3}{c}{\textit{Not passed}} & \multicolumn{3}{c}{\textit{Not passed}} & \multicolumn{3}{c}{\textit{Not passed}}\\
DeepSeek V4.1 Flash & \multicolumn{3}{c}{\textit{Not passed}} & \multicolumn{3}{c}{\textit{Not passed}} & \multicolumn{3}{c}{\textit{Not passed}}\\
\bottomrule
\end{tabular*}
\caption{Serving performance across three deployments. Entries pair
latency-qualified dollars per dollar with offered session concurrency $c$;
TTFT (s) and TPOT (ms) are p90 latencies. Shading marks the strongest baseline
and largest agent value in each~deployment.}
\label{tab:main-results}
\end{table*}

\subsection{System Verification}
\label{sec:validity}
After development, ServerBench rebuilds the submission in a fresh
deployment and replays sessions sampled with seeds disjoint from development. Independent tests and
post-run review check the resulting system (\Cref{fig:interface}).

\paragraph{Workload and output verification.}
Replay checks cover complete prompts, output lengths, streaming consistency,
and session progress. Timed replay disables EOS stopping to generate the prescribed output length;
separate EOS probes exercise natural stopping. Reference-scored
continuations check generation against the checkpoint. Held-out questions
compare answer correctness with a plain implementation of the same model,
using the task's accuracy tolerance to screen for substantial regressions
and shortcuts that bypass model computation.
Fresh deployments also test behavior beyond development state.

\paragraph{Post-run review.}
A review agent and human inspection examine submitted code and
recorded actions for checkpoint fidelity, inference semantics, and
workload execution. Review can identify prohibited shortcuts missed by
behavioral tests and overturn automated failures when inspection establishes
compliance with task requirements. These checks determine
score eligibility (\Cref{eq:objective}).

\keepspace{6\baselineskip}
\section{Evaluation}
\label{sec:evaluation}

\subsection{Experimental Setup}
\label{sec:setup}
We evaluate 11 coding-agent models (\Cref{tab:main-results}): Claude
\citep{anthropic2026opus5,anthropic2026opus55,anthropic2026sonnet55},
GPT \citep{openai2026gpt6}, Kimi \citep{kimi2026k3}, GLM
\citep{zai2026glm53}, DeepSeek \citep{deepseek2026v41flash}, MiMo
\citep{xiaomi2026mimo26pro}, and Qwen \citep{qwen2026max0902} models.
All use maximum reasoning effort on Qwen3.5 9B ($2\times$L40), Gemma-4-E4B-it
(Gemma E4B; $1\times$L40), and Qwen3.8 27B INT4 ($2\times$L40).
In Harbor, GPT models use Codex \citep{openai2026codex},
and other models use Claude Code \citep{anthropic2026claudecode}.
Each agent--deployment pair allows one 8 hours development run and
a 1,800-second final replay, with at most one rerun for a clearly anomalous
failed outcome.
We report each pair's strongest verified final submission across the
33 \Cref{tab:main-results} entries; development analysis focuses on
30 runs with server execution records.

Baselines are SGLang v0.5.19 and vLLM v0.28.0
\citep{zheng2024sglang,kwon2023efficient}. SGLang uses each model's
cookbook configuration; reference results average replay seeds at fixed
concurrency. Controlled comparisons replay frozen implementations and
intermediate versions on common seeds.

\subsection{End-to-End Results}
\label{sec:main-results}
\label{sec:value-analysis}
\Cref{tab:main-results} shows that agents can design verified serving
systems with higher value than the strongest reference on each deployment.
Claude Opus 5.5 leads Qwen3.5 9B and Qwen3.8 27B INT4 with 4.40 and 2.64
dollars per dollar, $2.71\times$ the SGLang reference and $1.80\times$ the
vLLM reference, respectively. Claude Sonnet 5.5 achieves 2.97 on Gemma
E4B, $4.87\times$ the vLLM reference.

Performance remains uneven across agents. Across 33 runs, 22 yield verified systems and 13
exceed the strongest reference. Six models deliver verified systems on
all three deployments, but only three exceed the strongest reference on
all three. GPT 6.1 Sol achieves 2.84 on Qwen3.5 9B, while its Qwen3.8 27B
INT4 system scores 0.98 against a reference of 1.47. Opus 5 passes verification but trails both Qwen references.

\begin{figure}[!t]
\centering
\includegraphics[width=\textwidth]{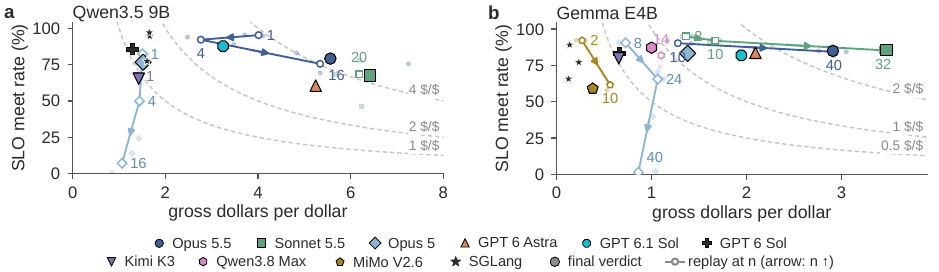}
\caption{Gross value $G$ versus value-weighted SLO meet rate $f$.
Large markers show final evaluations; open markers compare frozen
systems at different loads. Arrows indicate increasing
concurrency; dashed curves mark equal $D=Gf$.}
\label{fig:value-analysis}
\end{figure}

\Cref{fig:value-analysis} separates gross value from the SLO meet values. In the final replays
on Qwen3.5 9B, Sonnet 5.5 produces more gross value than Opus 5.5
(6.42 versus 5.56), but a smaller share qualifies, yielding 4.34 and 4.40. \Cref{fig:cost-value} contrasts GPT 6.1 Sol's lower-cost gains
with Claude's higher serving value. We next analyze agents' design,
feedback, and verification decisions.

\FloatBarrier
\subsection{High-Value Designs Coordinate Capacity and Concurrency}
\label{sec:loop-analysis}
\label{sec:analysis}
\label{sec:concurrency-analysis}
\begin{wrapfigure}[22]{r}{0.45\linewidth}
\centering
\includegraphics[width=0.94\linewidth]{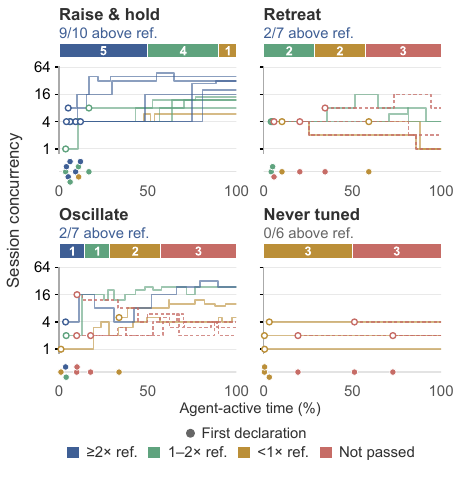}
\caption{Recorded concurrency paths and final outcomes. Colors mark value relative to
the strongest baseline; labels count runs. Dots below panels mark first
load declarations.}
\label{fig:decision-making}
\end{wrapfigure}
At fixed GPU cost, additional sessions can increase token value but raise
TTFT and TPOT through queueing and contention. Agents must match load
to serving capacity.
\Cref{fig:decision-making} relates recorded concurrency patterns to
final economic outcomes; runs that raise and hold load more often exceed
the strongest reference.
Among runs reaching at least twice the strongest reference, 7 of
17 load decisions improve value on subsequent measurement,
compared with 2 of 23 for verified runs below the reference
score.

Early end-to-end replay gives agents time to develop capacity while
adjusting load. The former group reaches usable replay in a median of
0.90 hours, compared with 2.96 hours for the latter. The optimization
inventory and controlled comparisons in
\Cref{app:capacity-comparisons} connect these operating decisions
to changes in execution, state management, and scheduling. Higher-value
systems tend to combine more serving optimizations, including speculative
decoding, host-memory state storage, and latency-aware scheduling.
Cross-turn reuse is common to all audited systems.

Controlled comparisons illustrate this distinction. At 16 sessions,
Opus 5.5's v1$\to$v2 Qwen redesign raises value by $1.32$--$1.88\times$
on matched seeds. For frozen Gemma systems, Sonnet 5.5 increases value
from 1.30 at two sessions to 2.97 at 32. Sonnet 5.5 also holds four Gemma sessions
while revising the engine, then increases load together with further
implementation changes. The trajectory illustrates capacity development before increasing load.
On Qwen, Opus 5 instead lowers concurrency
from four to two to one as it refines the implementation. Kimi likewise
reduces load to one after identifying long prefills that interfere with
decoding. Load reduction relieves contention; capacity improvements
support more simultaneous work.

Agents estimate the economic operating point in different ways. Opus 5.5
uses a workload simulator to guide Qwen load tests, then retains 16
sessions after a 20-session replay yields more gross value but only
64\% qualifying value. Sonnet 5.5 on Gemma compares settings on a fixed
seed: 36 sessions produce more gross value but incur more TPOT
misses, and it submits 32. These decisions match offered load to serving capacity within the latency targets.

\subsection{Using Feedback to Guide Optimization}
\label{sec:reward-analysis}
Agents use performance measurements and diagnostic tests to retain,
revise, or abandon changes. Useful feedback requires testing a change
under relevant workload conditions. Costly replays and seed variation
make diagnostic tests and workload analysis essential within the
development budget.

\begin{figure}[t]
\centering
\includegraphics[width=\textwidth]{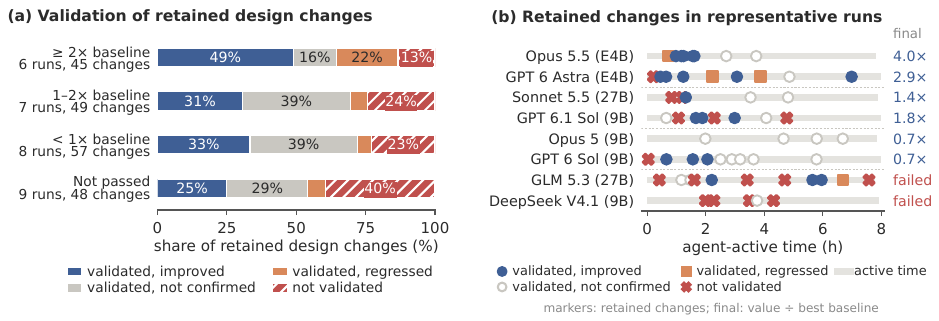}
\caption{Retained changes more often show workload-relevant gains
in the $\geq2\times$ group (a); representative trajectories show their
follow-up (b). Groups use final value relative to the strongest baseline.}
\label{fig:design-loop}
\end{figure}

\Cref{fig:design-loop} connects retained design decisions to
the tests supporting them. Among runs reaching at least twice the strongest
reference, 49\% of retained decisions follow improvements on
workload-relevant tests, versus 25\% in failed runs; decisions without
such tests account for 13\% and 40\%, respectively.
In the other verified groups, 39\% remain unconfirmed. These outcomes highlight testing and follow-up when deciding which changes to retain.

\suppressfloats[t]
On Gemma, experimental judgment changes whether agents pursue
prompt-lookup speculation. Opus 5 accepts a null result from a synthetic
random-text probe and omits the optimization. Opus 5.5 investigates its
initial null result using 60 replies from replay, repairs its
simulator, and measures 2.09--2.34 accepted tokens per step before adopting
speculation. At a fixed assumed draft cost, a corrected, length-matched
probe selects a smaller draft from synthetic text than from workload replies
(\Cref{app:finding-guided}).
To test the effect of this workload feedback on the agent's decision,
we resume Opus 5 from the same checkpoint for two hours per arm,
with or without feedback from its own replay replies.
Only the feedback arm re-adopts speculation;
its server delivers $1.13$--$1.31\times$ the qualified value on two fresh
common seeds. Workload feedback improves the agent's design decision and
its submitted server's value.

Favorable feedback can also lead agents to retain a configuration that
performs poorly under other workload conditions. In one Opus 5.5 run on Qwen3.8 27B INT4, the agent raises
concurrency from 10 to 16 and revises scheduling, observing a score
increase from 2.20 to 2.99. It recognizes overload and acknowledges that
changing seeds complicate comparison, but judges the additional value
worth the latency losses. Near the deadline, a scheduling repair remains
in warmup, so it returns to the configuration supported by one completed
900-second replay. In the final replay, the same implementation's
qualified share falls from 76.0\% to 16.8\% and its score to 0.48.

Workload coverage can be checked cheaply. Our CPU scan of the
workload generator over 128 additional seeds takes only 7.71 seconds:
the development draw lies near the fourth percentile of initial stream
availability, while the final draw is near the median. The agent had
inspected corpus statistics, yet used an unusually light initial branching
condition to validate its final configuration
(\Cref{app:controlled-measurement}).

Null and favorable results both require checking what the test exercises.
Inspecting test inputs and profiling workload variation helps agents
select representative and demanding conditions for costly replay,
supporting decisions to retain, revise, or abandon a design.

\subsection{Verification Must Match the Deployed Workload}
\label{sec:verification-analysis}
\begin{figure}[!t]
\centering
\includegraphics[width=\textwidth]{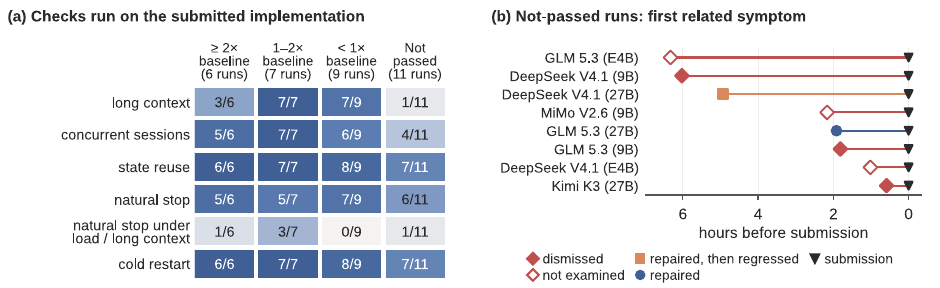}
\caption{Workload-relevant checks on submitted implementations (a).
Related symptoms precede submission in 8 of 11 not-passed runs (b);
markers distinguish responses and recorded submission times.}
\label{fig:verification}
\end{figure}

Verification checks the request boundaries and state transitions affected
by changes to admission, scheduling, and state reuse, including after
subsequent edits. \Cref{fig:verification} shows sparse coverage
at submission time: only 5 of the 33 entries have recorded checks of
natural stopping on the submitted system under load or long context.

Admission policies require checking which requests can make progress.
On Qwen3.5 9B, Qwen3.8-Max-0902 tightens admission to protect TPOT, then checks
a short prompt and a replay completing 11 requests. Its submitted
policy leaves a range of permitted prompt lengths unable to enter
execution even after other requests finish. Final evaluation stalls on
a 315K-token input in this range. The successful short tests leave the
new policy's prompt-length boundaries unchecked.

State reuse requires checking behavior across requests. On Qwen3.8 27B
INT4, Kimi's replay exposes a prefix-cache exception alongside a numerical
error. The agent repairs the numerical path and successfully retests an
isolated long prompt with a fresh engine. A separate check of the
submitted cache succeeds on the first insertion but crashes when the same
prefix is inserted again; the exception also recurs in final evaluation.
The isolated test misses prefix reuse across requests.

Opus 5.5 on Gemma explicitly tests the operating conditions made
relevant by its design. After adding preemption and cache restoration to coordinate TTFT
and TPOT, it reduces cache capacity to force these paths to execute.
The stress test exposes invalid outputs from uninitialized cache storage.
The defect predates the preemption change. The agent repairs cache access,
checks that unused slots filled with NaNs cannot affect valid computation,
and reruns forced preemption while comparing resumed outputs with isolated
execution. The final system retains these repairs. Reduced-pool replication
passes 0/3 tests with the repair reverted and 3/3 after repair.

{\parfillskip=0pt plus 1fil 
Reliable delivery requires testing admission boundaries, shared state,
and recovery, then rerunning affected checks after~repairs.\par}

\subsection{Design and Workload Ablations}
\label{sec:ablation}
Finding-guided changes improve serving value and correctness
(\Cref{app:finding-guided}). On two fresh common seeds under the
final 1,800-second protocol, Opus 5.5's Qwen v1$\to$v6 redesign at $c=16$
raises qualified value by $1.35\times$ and $1.15\times$.
On frozen Gemma at $c=40$, workload-selected draft length 7 outperforms
synthetic-selected length 3 by 3.0\% and 3.9\%.
For workload-matched verification, Opus 5.5's Gemma cache repair
passes all 3 reduced-pool stress tests
(\Cref{sec:verification-analysis}).

\begin{figure}[!t]
\centering
\setlength{\abovecaptionskip}{4pt}
\includegraphics[width=\textwidth]{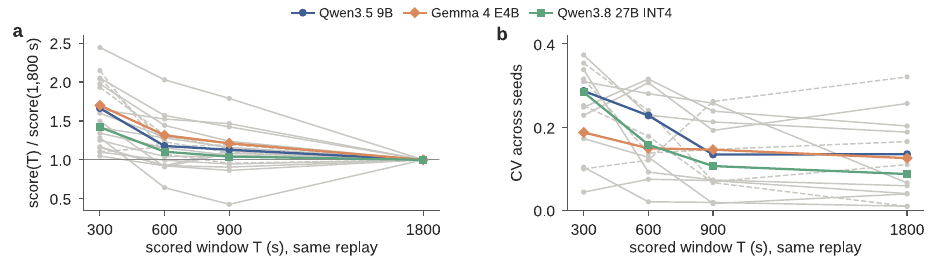}
\caption{Short windows yield higher average value (a); sampling
sensitivity varies across systems (b). Colored curves average systems
within each deployment.}
\label{fig:measurement-conditions}
\end{figure}

\Cref{fig:measurement-conditions} fixes implementation, concurrency,
and sessions. Across deployments, 300-second windows average
$1.42$--$1.70\times$ the 1,800-second scores, versus
$1.04$--$1.21\times$ at 900 seconds. For Opus 5.5 on Qwen,
extending the window from 300 to 1,800 seconds lowers the across-seed
coefficient of variation from 0.37 to 0.19, yielding more stable
performance estimates.
The closer agreement at 900 seconds supports using intermediate replays
to guide development. Remaining variation motivates combining these
measurements with workload profiling and targeted diagnostics to select
configurations and workload conditions for longer replay.

\FloatBarrier
\section{Conclusion}
ServerBench evaluates agentic design of complete, specialized LLM serving
systems through an economic service objective, stateful closed-loop
replay, and independent verification. Results across three deployments
show substantial gains alongside uneven delivery. Trajectory analysis
and controlled comparisons highlight three priorities: coordinating
capacity improvements with operating load, designing experiments around
workload characteristics and variation, and verifying the execution paths
introduced by system changes.

\section*{Limitations}
Our evaluation covers three model--hardware deployments and one coding-trace
source. Development and final draws share a corpus; reconstructed inputs
preserve workload structure without reproducing original coding tasks.
Closed-loop replay differs from open-loop production demand, limiting
generalization to other architectures, accelerators, and arrival patterns.

The comparisons characterize measured artifacts. Correlations, source inspection,
and selected trajectories do not establish causal optimization effects or
design success rates. Repeated agent runs and replays are needed to
separate agent variability from workload variability. Version notes record
reported design changes and can omit implementation details. Behavioral tests
and code review have finite coverage, and calibration depends on the reference
engine. Fixed-price token value measures deployment efficiency, excluding
development and other operating costs, rather than realized profit.

\section*{Ethical Considerations}
Recorded coding sessions can contain sensitive information, and
agent-generated serving code can introduce unsafe behavior. ServerBench
replaces original session content during workload reconstruction and
executes generated code in isolated environments with restricted network
access. Any release of traces or agent trajectories should be reviewed
for sensitive information and comply with source licenses. Post-run code
and action review supports the identification of unauthorized behavior.
Controlled GPU allocations and experiment resource reporting support
responsible use. These controls reduce exposure but do not establish that
arbitrary generated code is safe.

\section*{Acknowledgements}
This work was partially supported by Google Research Award, Google ML \& System Junior Faculty Award, Amazon Research Award, Fireworks AI, Intel, Li Auto, Moffett AI, Daytona, and CMU CyLab Seed funding. This material is also based upon work supported by the National Science Foundation under Grant Nos. CCF-2504353 and CCF-2247014, and by IARPA. Any opinions, findings, conclusions or recommendations expressed are those of the authors and do not necessarily reflect the views of the National Science Foundation.

\clearpage

\bibliographystyle{assets/plainnat}
\bibliography{references}

\begin{thebibliography}{40}
\providecommand{\natexlab}[1]{#1}
\providecommand{\url}[1]{\texttt{#1}}
\expandafter\ifx\csname urlstyle\endcsname\relax
  \providecommand{\doi}[1]{doi: #1}\else
  \providecommand{\doi}{doi: \begingroup \urlstyle{rm}\Url}\fi

\bibitem[Agrawal et~al.(2024)Agrawal, Kedia, Panwar, Mohan, Kwatra, Gulavani, Tumanov, and Ramjee]{agrawal2024taming}
Amey Agrawal, Nitin Kedia, Ashish Panwar, Jayashree Mohan, Nipun Kwatra, Bhargav Gulavani, Alexey Tumanov, and Ramachandran Ramjee.
\newblock Taming throughput-latency tradeoff in {LLM} inference with {Sarathi-Serve}.
\newblock In \emph{18th USENIX symposium on operating systems design and implementation (OSDI 24)}, pages 117--134, 2024.

\bibitem[{Anthropic}(2026{\natexlab{a}})]{anthropic2026claudecode}
{Anthropic}.
\newblock {Claude Code}, 2026{\natexlab{a}}.
\newblock \url{https://github.com/anthropics/claude-code}.
\newblock Software repository. Accessed October 12, 2026.

\bibitem[{Anthropic}(2026{\natexlab{b}})]{anthropic2026opus5}
{Anthropic}.
\newblock Introducing {Claude Opus 5}, 2026{\natexlab{b}}.
\newblock \url{https://www.anthropic.com/news/claude-opus-5}.

\bibitem[{Anthropic}(2026{\natexlab{c}})]{anthropic2026opus55}
{Anthropic}.
\newblock Introducing {Claude Opus 5.5}, 2026{\natexlab{c}}.
\newblock \url{https://www.anthropic.com/claude-opus-5-5}.

\bibitem[{Anthropic}(2026{\natexlab{d}})]{anthropic2026sonnet55}
{Anthropic}.
\newblock Introducing {Claude Sonnet 5.5}, 2026{\natexlab{d}}.
\newblock \url{https://www.anthropic.com/claude-sonnet-5-5}.

\bibitem[Chan et~al.(2025)Chan, Chowdhury, Jaffe, Aung, Sherburn, Mays, Starace, Liu, Maksin, Patwardhan, et~al.]{chan2025mle}
Jun~Shern Chan, Neil Chowdhury, Oliver Jaffe, James Aung, Dane Sherburn, Evan Mays, Giulio Starace, Kevin Liu, Leon Maksin, Tejal Patwardhan, et~al.
\newblock {MLE-bench}: Evaluating machine learning agents on machine learning engineering.
\newblock In \emph{International Conference on Learning Representations}, volume 2025, pages 50466--50494, 2025.

\bibitem[Chen et~al.(2026)Chen, Zhong, Feng, Sadhukhan, Zhou, Shieh, Jia, and Chen]{chen2026vortex}
Zhuoming Chen, Xinrui Zhong, Qilong Feng, Ranajoy Sadhukhan, Yang Zhou, Michael~Qizhe Shieh, Zhihao Jia, and Beidi Chen.
\newblock Vortex: Efficient and programmable sparse attention serving for {AI} agents.
\newblock \emph{arXiv preprint arXiv:2606.06453}, 2026.

\bibitem[Cheng et~al.(2025)Cheng, Liu, Pan, Li, Wang, Krentsel, Xia, Cemri, Park, Yang, et~al.]{cheng2025barbarians}
Audrey Cheng, Shu Liu, Melissa Pan, Zhifei Li, Bowen Wang, Alex Krentsel, Tian Xia, Mert Cemri, Jongseok Park, Shuo Yang, et~al.
\newblock Barbarians at the gate: How ai is upending systems research.
\newblock \emph{arXiv preprint arXiv:2510.06189}, 2025.

\bibitem[{DeepSeek-AI}(2026)]{deepseek2026v41flash}
{DeepSeek-AI}.
\newblock {DeepSeek-V4.1-Flash}: Pushing the limits of {KV} cache compression.
\newblock \emph{arXiv preprint arXiv:2609.19969}, 2026.
\newblock \url{https://arxiv.org/abs/2609.19969}.

\bibitem[Ding et~al.(2026)Ding, Wang, Huang, Wan, Zhang, Han, Xu, Zhang, Chu, You, et~al.]{ding2026phi}
Leilei Ding, Shumin Wang, Yuting Huang, Fanqi Wan, Yinmin Zhang, Qi~Han, Yiming Xu, Feiyuan Zhang, Xiaomeng Chu, Guoliang You, et~al.
\newblock $\phi$-bench: Can large language models engineer the infrastructure that powers them?
\newblock \emph{arXiv preprint arXiv:2609.10226}, 2026.

\bibitem[{Google}(2026)]{google2026gemma4e4b}
{Google}.
\newblock {Gemma-4-E4B-it}, 2026.
\newblock \url{https://huggingface.co/google/gemma-4-E4B-it}.
\newblock Model card.

\bibitem[{Kimi Team}(2026)]{kimi2026k3}
{Kimi Team}.
\newblock {Kimi K3}: Open frontier intelligence.
\newblock \emph{arXiv preprint arXiv:2607.24653}, 2026.
\newblock \url{https://arxiv.org/abs/2607.24653}.

\bibitem[Kwon et~al.(2023)Kwon, Li, Zhuang, Sheng, Zheng, Yu, Gonzalez, Zhang, and Stoica]{kwon2023efficient}
Woosuk Kwon, Zhuohan Li, Siyuan Zhuang, Ying Sheng, Lianmin Zheng, Cody~Hao Yu, Joseph Gonzalez, Hao Zhang, and Ion Stoica.
\newblock Efficient memory management for large language model serving with {PagedAttention}.
\newblock In \emph{Proceedings of the 29th symposium on operating systems principles}, pages 611--626, 2023.

\bibitem[Merrill et~al.(2026)Merrill, Shaw, Carlini, Li, Raj, Bercovich, Shi, Shin, Walshe, Buchanan, et~al.]{merrill2026terminal}
Mike Merrill, Alexander Shaw, Nicholas Carlini, Boxuan Li, Harsh Raj, Ivan Bercovich, Lin Shi, Jeong Shin, Thomas Walshe, E~Kelly Buchanan, et~al.
\newblock {Terminal-Bench}: Benchmarking agents on hard, realistic tasks in command line interfaces.
\newblock In \emph{International Conference on Learning Representations}, volume 2026, pages 40903--40986, 2026.

\bibitem[Nangia et~al.(2026)Nangia, Mishra, Gokrani, and Chopra]{nangia2026iso}
Ayush Nangia, Shikhar Mishra, Aman Gokrani, and Paras Chopra.
\newblock {ISO-Bench}: Can coding agents optimize real-world inference workloads?
\newblock \emph{arXiv preprint arXiv:2602.19594}, 2026.

\bibitem[Novikov et~al.(2025)Novikov, V{\~u}, Eisenberger, Dupont, Huang, Wagner, Shirobokov, Kozlovskii, Ruiz, Mehrabian, et~al.]{novikov2025alphaevolve}
Alexander Novikov, Ng{\^a}n V{\~u}, Marvin Eisenberger, Emilien Dupont, Po-Sen Huang, Adam~Zsolt Wagner, Sergey Shirobokov, Borislav Kozlovskii, Francisco~JR Ruiz, Abbas Mehrabian, et~al.
\newblock Alphaevolve: A coding agent for scientific and algorithmic discovery.
\newblock \emph{arXiv preprint arXiv:2506.13131}, 2025.

\bibitem[{OpenAI}(2026{\natexlab{a}})]{openai2026codex}
{OpenAI}.
\newblock {Codex}, 2026{\natexlab{a}}.
\newblock \url{https://github.com/openai/codex}.
\newblock Software repository. Accessed October 12, 2026.

\bibitem[{OpenAI}(2026{\natexlab{b}})]{openai2026gpt6}
{OpenAI}.
\newblock Using {GPT-6}, 2026{\natexlab{b}}.
\newblock \url{https://developers.openai.com/api/docs/guides/latest-model}.
\newblock API documentation. Accessed October 11, 2026.

\bibitem[Ouyang et~al.(2025)Ouyang, Guo, Arora, Zhang, Hu, R{\'e}, and Mirhoseini]{ouyang2025kernelbench}
Anne Ouyang, Simon Guo, Simran Arora, Alex~L Zhang, William Hu, Christopher R{\'e}, and Azalia Mirhoseini.
\newblock {KernelBench}: Can {LLMs} write efficient {GPU} kernels?
\newblock \emph{arXiv preprint arXiv:2502.10517}, 2025.

\bibitem[{Qwen Team}(2026{\natexlab{a}})]{qwen2026max0902}
{Qwen Team}.
\newblock {Qwen3.8-Max-0902}, 2026{\natexlab{a}}.
\newblock \url{https://www.qwencloud.com/models/qwen3.8-max-0902}.
\newblock Official model page. Accessed October 11, 2026.

\bibitem[{Qwen Team}(2026{\natexlab{b}})]{qwen2026qwen35_9b}
{Qwen Team}.
\newblock {Qwen3.5-9B}, 2026{\natexlab{b}}.
\newblock \url{https://huggingface.co/Qwen/Qwen3.5-9B}.
\newblock Model card.

\bibitem[{Red Hat AI}(2026)]{redhatai2026qwen38_27b_int4}
{Red Hat AI}.
\newblock {Qwen3.8-27B-INT4}, 2026.
\newblock \url{https://huggingface.co/RedHatAI/Qwen3.8-27B-INT4}.
\newblock Model card.

\bibitem[Reddi et~al.(2020)Reddi, Cheng, Kanter, Mattson, Schmuelling, Wu, Anderson, Breughe, Charlebois, Chou, et~al.]{reddi2020mlperf}
Vijay~Janapa Reddi, Christine Cheng, David Kanter, Peter Mattson, Guenther Schmuelling, Carole-Jean Wu, Brian Anderson, Maximilien Breughe, Mark Charlebois, William Chou, et~al.
\newblock {MLPerf} inference benchmark.
\newblock In \emph{2020 ACM/IEEE 47th Annual International Symposium on Computer Architecture (ISCA)}, pages 446--459. IEEE, 2020.

\bibitem[{SemiAnalysis}(2026{\natexlab{a}})]{semianalysis2026agentx}
{SemiAnalysis}.
\newblock {AgentX Harness}.
\newblock \url{https://github.com/SemiAnalysisAI/agentx-harness}, 2026{\natexlab{a}}.
\newblock Open-source software.

\bibitem[{SemiAnalysis}(2026{\natexlab{b}})]{semianalysis2026agentxdata}
{SemiAnalysis}.
\newblock {AgentX} methodology and datasets, 2026{\natexlab{b}}.
\newblock \url{https://inferencex.semianalysis.com/agentx}.
\newblock Dataset and methodology. Accessed October 12, 2026.

\bibitem[Srivatsa et~al.(2025)Srivatsa, He, Abhyankar, Li, and Zhang]{srivatsa2025preble}
Vikranth Srivatsa, Zijian He, Reyna Abhyankar, Dongming Li, and Yiying Zhang.
\newblock {Preble}: Efficient distributed prompt scheduling for {LLM} serving.
\newblock In \emph{International conference on learning representations}, volume 2025, pages 37057--37082, 2025.

\bibitem[Sun et~al.(2026)Sun, Han, Zhang, Pang, Wang, Cao, Huang, Duroiu, Zhang, Lin, et~al.]{sun2026ale}
Yiyou Sun, Xinyang Han, Weichen Zhang, Yuanbo Pang, Tianyu Wang, Yuhan Cao, Yixiao Huang, Chris Duroiu, Haoyun Zhang, Jeffrey Lin, et~al.
\newblock Agents' last exam.
\newblock \emph{arXiv preprint arXiv:2606.05405}, 2026.
\newblock \url{https://arxiv.org/abs/2606.05405}.

\bibitem[Williams et~al.(2026)Williams, Farris, Farris, and Jiao]{williams2026swe}
Jennifer Williams, Dave Farris, Jeff Farris, and Jiantao Jiao.
\newblock {SWE-Serve}: Benchmarking agentic engineering for production inference serving.
\newblock \emph{arXiv preprint arXiv:2609.26777}, 2026.

\bibitem[Xiao et~al.(2025)Xiao, Wang, Kong, Golac, and Wang]{xiao2025csr}
Yijia Xiao, Runhui Wang, Luyang Kong, Davor Golac, and Wei Wang.
\newblock {CSR-Bench}: Benchmarking {LLM} agents in deployment of computer science research repositories.
\newblock In \emph{Proceedings of the 2025 Conference of the Nations of the Americas Chapter of the Association for Computational Linguistics: Human Language Technologies (Volume 1: Long Papers)}, pages 12705--12723, 2025.

\bibitem[{Xiaomi MiMo}(2026)]{xiaomi2026mimo26pro}
{Xiaomi MiMo}.
\newblock {MiMo-V2.6-Pro}, 2026.
\newblock \url{https://mimo.mi.com/models/en-US/mimo-v2.6-pro}.
\newblock Official model page. Accessed October 11, 2026.

\bibitem[Xu et~al.(2026)Xu, Liu, Li, Rabbani, and Zhang]{xu2026seis}
Zhen Xu, Jingyu Liu, Zongze Li, Tahseen Rabbani, and Ce~Zhang.
\newblock {SEIS}: Self-evolving inference systems.
\newblock \emph{arXiv preprint arXiv:2610.04646}, 2026.

\bibitem[Yang et~al.(2024)Yang, Jimenez, Wettig, Lieret, Yao, Narasimhan, and Press]{yang2024swe}
John Yang, Carlos Jimenez, Alexander Wettig, Kilian Lieret, Shunyu Yao, Karthik Narasimhan, and Ofir Press.
\newblock {SWE-agent}: Agent-computer interfaces enable automated software engineering.
\newblock \emph{Advances in Neural Information Processing Systems}, 37:\penalty0 50528--50652, 2024.

\bibitem[Yeon et~al.(2026)Yeon, Rank, and Andriushchenko]{yeon2026inferencebench}
Jehyeok Yeon, Ben Rank, and Maksym Andriushchenko.
\newblock {InferenceBench}: A benchmark for open-ended {LLM} inference optimization by {AI} agents.
\newblock \emph{arXiv preprint arXiv:2607.20468}, 2026.

\bibitem[Yu et~al.(2022)Yu, Jeong, Kim, Kim, and Chun]{yu2022orca}
Gyeong-In Yu, Joo~Seong Jeong, Geon-Woo Kim, Soojeong Kim, and Byung-Gon Chun.
\newblock {Orca}: A distributed serving system for {Transformer}-based generative models.
\newblock In \emph{16th USENIX symposium on operating systems design and implementation (OSDI 22)}, pages 521--538, 2022.

\bibitem[{Z.ai}(2026)]{zai2026glm53}
{Z.ai}.
\newblock {GLM-5.3}, 2026.
\newblock \url{https://docs.z.ai/guides/llm/glm-5.3}.
\newblock Developer documentation. Accessed October 11, 2026.

\bibitem[Zheng et~al.(2024)Zheng, Yin, Xie, Sun, Huang, Yu, Cao, Kozyrakis, Stoica, Gonzalez, et~al.]{zheng2024sglang}
Lianmin Zheng, Liangsheng Yin, Zhiqiang Xie, Chuyue Sun, Jeff Huang, Cody~H Yu, Shiyi Cao, Christos Kozyrakis, Ion Stoica, Joseph~E Gonzalez, et~al.
\newblock {SGLang}: Efficient execution of structured language model programs.
\newblock \emph{Advances in neural information processing systems}, 37:\penalty0 62557--62583, 2024.

\bibitem[Zheng et~al.(2025)Zheng, Cheng, Shen, Zhou, Liu, He, Li, Wei, Hao, Yao, Sheng, Wang, Chai, Korolova, Henderson, Arora, Viswanath, Shang, and Xie]{zheng2025livecodebenchpro}
Zihan Zheng, Zerui Cheng, Zeyu Shen, Shang Zhou, Kaiyuan Liu, Hansen He, Dongruixuan Li, Stanley Wei, Hangyi Hao, Jianzhu Yao, Peiyao Sheng, Zixuan Wang, Wenhao Chai, Aleksandra Korolova, Peter Henderson, Sanjeev Arora, Pramod Viswanath, Jingbo Shang, and Saining Xie.
\newblock {LiveCodeBench Pro}: How do olympiad medalists judge {LLMs} in competitive programming?
\newblock In \emph{Advances in Neural Information Processing Systems}, volume~38, 2025.
\newblock \url{https://proceedings.neurips.cc/paper_files/paper/2025/hash/9712b78386cebdc3db7f1a48c2d20edb-Abstract-Datasets_and_Benchmarks_Track.html}.

\bibitem[Zhong et~al.(2024)Zhong, Liu, Chen, Hu, Zhu, Liu, Jin, and Zhang]{zhong2024distserve}
Yinmin Zhong, Shengyu Liu, Junda Chen, Jianbo Hu, Yibo Zhu, Xuanzhe Liu, Xin Jin, and Hao Zhang.
\newblock {DistServe}: Disaggregating prefill and decoding for goodput-optimized large language model serving.
\newblock In \emph{18th USENIX symposium on operating systems design and implementation (OSDI 24)}, pages 193--210, 2024.

\bibitem[Zhou et~al.(2026{\natexlab{a}})Zhou, Zheng, Liu, Shen, Cheng, Chen, He, Yao, Mao, Mang, Fu, Li, Li, Chai, Liu, Korolova, Henderson, Jaques, Viswanath, Xie, and Shang]{zhou2026autocode}
Shang Zhou, Zihan Zheng, Kaiyuan Liu, Zeyu Shen, Zerui Cheng, Zexing Chen, Hansen He, Jianzhu Yao, Huanzhi Mao, Qiuyang Mang, Tianfu Fu, Beichen Li, Dongruixuan Li, Wenhao Chai, Zhuang Liu, Aleksandra Korolova, Peter Henderson, Natasha Jaques, Pramod Viswanath, Saining Xie, and Jingbo Shang.
\newblock {AutoCode}: {LLMs} as problem setters for competitive programming.
\newblock In \emph{International Conference on Learning Representations}, 2026{\natexlab{a}}.
\newblock \url{https://proceedings.iclr.cc/paper_files/paper/2026/hash/1bd78da0966968409e65b5e0b5dbb57e-Abstract-Conference.html}.

\bibitem[Zhou et~al.(2026{\natexlab{b}})Zhou, Wang, Zhou, Liu, Wang, Peng, Wang, Zhang, Yin, Tian, Fu, Liu, Peng, Yang, Cui, Miao, and Zhang]{zhou2026janus}
Tianbao Zhou, Yi~Wang, Yu~Zhou, Zirui Liu, Zhiming Wang, Yebo Peng, Yongfu Wang, Yi~Zhang, Jinrun Yin, Kemeng Tian, Fangcheng Fu, Tongxuan Liu, Tao Peng, Tong Yang, Bin Cui, Xupeng Miao, and Ke~Zhang.
\newblock {Janus}: Multi-{LLM} serving at production scale.
\newblock In \emph{Proceedings of the ACM SIGOPS 32nd Symposium on Operating Systems Principles}, 2026{\natexlab{b}}.
\newblock \doi{10.1145/3830418.3843867}.

\end{thebibliography}

\clearpage

\beginappendix
\suppressfloats[t]   
\begin{figure*}[t]
\centering
\includegraphics[width=\textwidth]{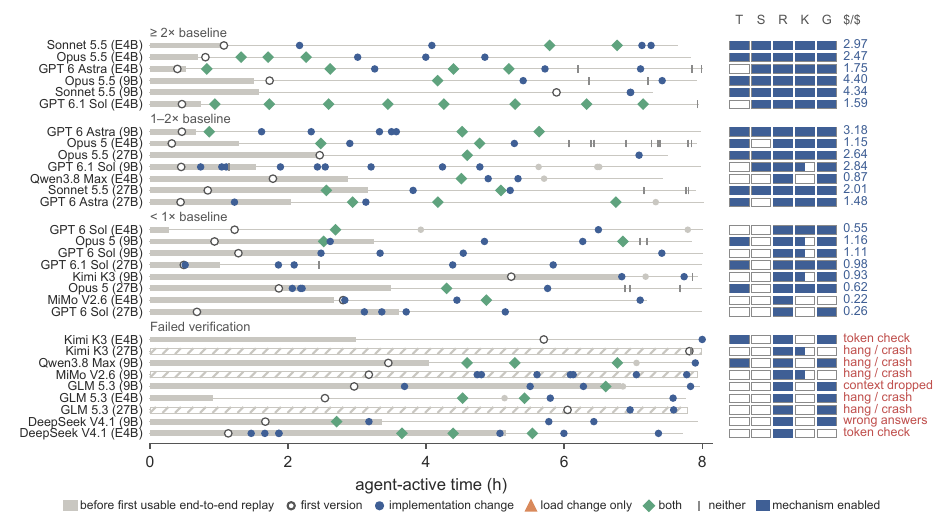}
\caption{Development trajectories connect published implementation changes to
the optimizations identified in submitted systems. Filled cells mark
latency-aware scheduling (T), speculative decoding (S), cross-turn reuse
(R), host-memory state storage (K), and CUDA graphs (G); half-filled K
cells indicate recurrent-state storage only. Groups use value relative to
each deployment's strongest reference; outcomes show dollars per dollar
or the failed check.}
\label{fig:design-inventory}
\end{figure*}

\section{Task Calibration and Accounting}
\label{app:calibration}
\Cref{tab:deployment-settings} summarizes the frozen deployment settings.

\begin{table*}[t]
\centering
\fontsize{9}{11}\selectfont
\setlength{\tabcolsep}{5pt}
\renewcommand{\arraystretch}{1.12}
\begin{tabular*}{\textwidth}{@{\extracolsep{\fill}}lccc@{}}
\toprule
Parameter & Qwen3.5 9B & Gemma E4B & Qwen3.8 27B INT4 \\
Checkpoint source & \citep{qwen2026qwen35_9b} & \citep{google2026gemma4e4b} & \citep{redhatai2026qwen38_27b_int4} \\
\midrule
GPU allocation & $2\times$L40 & $1\times$L40 & $2\times$L40 \\
Context limit (tokens) & 1,048,576 & 131,072 & 1,048,576 \\
Input price (USD / million tokens) & 0.1044 & 0.0200 & 0.0898 \\
Output price (USD / million tokens) & 0.1855 & 0.1000 & 2.6570 \\
TTFT target (s) & 4.10028 & 2.52693 & 10.2988 \\
TPOT target (ms) & 23.4974 & 121.7057 & 48.3769 \\
GPU cost (USD / GPU-hour) & 0.82 & 0.82 & 0.82 \\
\midrule
SGLang reference concurrency & 1 & 2 & 4 \\
vLLM reference concurrency & 1 & 8 & 1 \\
\bottomrule
\end{tabular*}
\caption{Deployment parameters, service targets, and reference operating points.}
\label{tab:deployment-settings}
\end{table*}
We calibrate SGLang and vLLM by selecting the concurrency that maximizes
gross dollars per dollar. Using 5 independent random seeds, we set each
latency target to the largest per-replay p90 to decide the SLO standard, and use the mean p90 TTFT, TPOT and dollars per dollar as the baseline. These targets are frozen before agent development and applied to individual requests; agents
choose their own operating concurrency. Changes to the model, hardware,
workload, or inference policy require recalibration under the revised
deployment conditions.

Reference results average 5 independent replay seeds at the operating
points in \Cref{tab:deployment-settings}.

Scores use the measured window duration and allocated GPU cost.
Cached input is valued at the full input-token price and counted once
per logical request. Value is computed from client-side token counts.
Completed requests contribute to the numerator of
\Cref{eq:objective} when both latency measurements are finite
and meet their respective targets.

\section{Replay and Reproducibility Record}
\label{app:reproducibility}
The replay uses SemiAnalysis's AgentX  \citep{semianalysis2026agentxdata} \nolinkurl{semianalysisai/cc-traces-weka-062126} coding sessions, with the corpus revision fixed
by the task. It samples session progress under a closed-loop load.
Development and final evaluation use disjoint
seed sets drawn from the same source corpus.

\paragraph{Workload reconstruction and eligibility.}
Deterministic reconstruction replaces original prompt and tool contents
while preserving request lengths, shared prefixes, and session topology.
Replay uses recorded end-to-start delays and session dependencies,
including parallel subagent branches. When no requests are in flight,
global idle gaps are capped at 10 seconds by moving all pending timed
events earlier by the same amount, preserving their relative spacing, order,
and dependency constraints. Context eligibility is fixed before development; where necessary,
whole traces are filtered to preserve dependencies without allowing
candidates to truncate prompts. State reuse must preserve all required
states, including recurrent state where applicable.

\paragraph{Development and deployment.}
The Transformers starter uses a global request lock and has no cross-request
prefix cache. Agents may change scheduling, batching, memory management,
kernels, parallel execution, and state reuse. Reference configurations
support comparison without prescribing the candidate architecture.
Harbor supports container or Modal backends; evaluation components remain
outside the editable server. Full development tests launch frozen snapshots,
while lightweight checks support shorter experiments. Publishing a version
and measuring its performance are separate operations. Portability checks
test compatibility with the deployment interface. After the final workload
replay, protocol and output tests run while the server remains active.
Final evaluation uses the frozen submission after the development budget ends.

\paragraph{Reproduction record.}
Reproduction requires the effective checkpoint and inference configuration,
task and corpus revisions, environment snapshot, agent model identifier and
scaffold, development budget, submitted artifact, chosen
concurrency, and replay seeds. Immutable version records connect intermediate
measurements to code; final results are associated with the submitted artifact.
The record separates construction runs from repeated replays of each
frozen system.

\section{Output Verification Settings}
\label{app:correctness}
The answer test samples 20 held-out questions and permits an absolute
accuracy decrease of at most 0.1 relative to a plain implementation of the
same checkpoint. For $n$ questions and $k_{\mathrm{ref}}$ correct reference
answers, the candidate must satisfy
\begin{equation}
 k_{\mathrm{cand}}\geq
 \max\{0,k_{\mathrm{ref}}-\lfloor0.1n\rfloor\}.
\end{equation}
The host runs this test on the frozen submission after development.

Replay checks compare native token identifiers with streamed text and
client accounting, and use context probes to test required extended-context
behavior. Candidates may not count malformed responses as completed work.
These checks assess faithful workload execution and token accounting.

Natural stopping is tested with end-of-sequence termination enabled.
The replay requests fixed output lengths, but tokens must still arise from
the checkpoint's computation. Suspicious continuations, including repeated
tokens and continuations after an end-of-sequence token, are examined using
reference scoring. Up to 16 suspect continuations are sampled, with a
maximum per-token log-probability regret tolerance of 4.0 under teacher forcing.
Timed replay disables end-of-sequence stopping to obtain prescribed output
lengths, but fabricated padding is prohibited. Continuations produced by
the prescribed model, including repetition, are permitted.

\section{Integrity Review and Agent Configuration}
\label{app:agent-configuration}
The task prompt fixes model semantics and prohibits reducing work through
truncation, approximate computation, precision reduction, answer retrieval,
or model substitution. It also prohibits identifying evaluation probes to
alter behavior, manipulating token or timing records, and modifying trusted
evaluation components. Computed-state reuse that preserves the specified
semantics is admissible.

The prompt announces post-run review of code and recorded actions by a
separate agent. The review examines the submitted implementation,
version changes, commands, and measurement records. Automated checks
record reported model identifiers, network destinations, and
commands indicating potential undeclared model use. These records support
code inspection alongside the final execution checks.
Human review complements these automated assessments when evaluating
suspected correctness failures and prohibited shortcuts.
Only maximum-reasoning-effort configurations appear in the main comparison; full execution
identifiers belong in the campaign reproducibility record.

\raggedbottom
\section{Development Analysis and Controlled Comparisons}
\label{app:additional-results}

\paragraph{Figure labels.}
Figure labels shorten Qwen3.8-Max-0902, DeepSeek V4.1 Flash, and MiMo V2.6 Pro
to Qwen3.8 Max, DeepSeek V4.1, and MiMo V2.6.

\paragraph{Development resources.}
Agent token consumption sums uncached input, cache reads, cache writes,
and output, including reasoning, across a run's continuation segments.
API cost estimates characterize development expenditure; serving costs
are accounted for by allocated GPU time.

\subsection{Agent API Costs}
\label{app:api-costs}
\Cref{tab:api-costs} reprices recorded token usage at OpenRouter's
advertised standard rates, retrieved on October 13, 2026.\footnote{\url{https://openrouter.ai/api/v1/models}}
We sum uncached input, cache reads, cache writes, and output, including
reasoning tokens once. Cache writes use the listed 5-minute rate, or the
ordinary input rate when no separate write price is listed. GPT prompts
remain below the 272K-token pricing threshold. The estimates cover
token and cache-operation charges.

\begin{table}[ht]
\centering
\fontsize{9}{11}\selectfont
\setlength{\tabcolsep}{4pt}
\begin{tabular*}{\columnwidth}{@{\extracolsep{\fill}}lrr@{}}
\toprule
Agent & Qwen 9B & Gemma E4B \\
\midrule
Claude Opus 5.5 & 99.69 & 104.25 \\
Claude Sonnet 5.5 & 78.60 & 30.72 \\
GPT 6 Astra & 110.81 & 119.79 \\
GPT 6.1 Sol & 13.01 & 13.67 \\
Claude Opus 5 & 108.90 & 104.37 \\
GPT 6 Sol & 50.88 & 35.93 \\
Kimi K3 & 126.98 & 28.71 \\
Qwen3.8-Max-0902 & 46.13 & 43.33 \\
MiMo V2.6 Pro & 6.32 & 10.19 \\
GLM 5.3 & 16.80 & 10.79 \\
DeepSeek V4.1 Flash & 1.64 & 2.07 \\
\bottomrule
\end{tabular*}
\caption{Agent API cost estimates (USD) at OpenRouter rates for \Cref{fig:cost-value}.}
\label{tab:api-costs}
\end{table}

\FloatBarrier
\keepspace{0.70\textheight}
\subsection{Design Changes and Decision Measurements}
\label{app:design-measurements}
\begin{figure}[h]
\centering
\includegraphics[width=\textwidth]{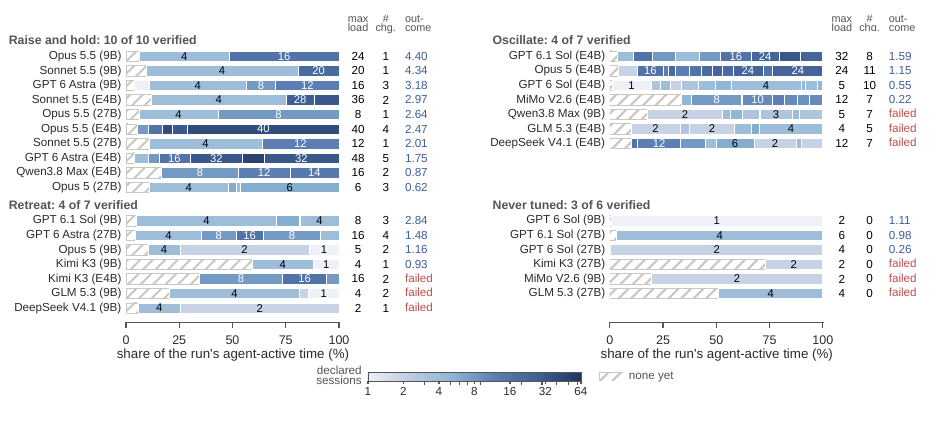}
\caption{Individual concurrency trajectories underlying
\Cref{fig:decision-making}. Rows show declared concurrency,
the maximum tested load, the number of changes, and the final outcome.}
\label{fig:concurrency-trajectories}
\end{figure}

\FloatBarrier
\begin{wrapfigure}{r}{0.45\linewidth}
\centering
\includegraphics[width=\linewidth]{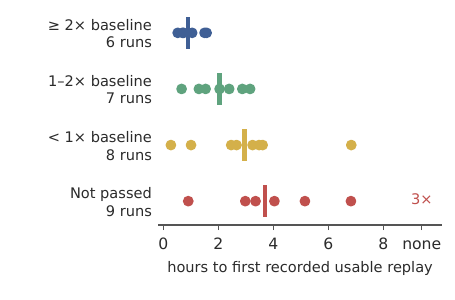}
\caption{Higher-value runs establish end-to-end replay earlier, leaving more
time for system development. Three failed runs never reach a usable replay.}
\label{fig:first-replay}
\end{wrapfigure}
The reconstruction connects published versions to their implementation,
declared concurrency, measurements, and recorded agent actions.
\Cref{fig:design-inventory} distinguishes implementation changes,
load-only changes, coordinated changes, and publications changing neither.
The first usable end-to-end measurement requires a non-mock, agent-owned
server, replay fidelity, and positive measured value. Implementation
inspection records
mechanisms enabled on the served path, including scheduling, speculation,
cross-turn reuse, host-memory state storage, and CUDA graphs. These
inventories show that higher-value systems tend to combine a broader set
of serving optimizations, particularly speculative decoding,
latency-aware scheduling, and host-memory state storage. Cross-turn reuse
is present across all audited systems, while CUDA graphs are also widespread.
Decision-episode labels record
the triggering signal, hypothesis, workload relevance, comparability,
action, follow-up, and support from the recorded comparison. The labeling scheme distinguishes
supported, partially supported, unsupported, and untestable conclusions
using the decision's tests and follow-up. Episodes cover tested hypotheses, rollback,
retention, and load selection; load changes are linked to their preceding signals.
Declared concurrency and the largest experimentally tested load are
recorded separately, and trajectories use agent-active time.

\Cref{fig:concurrency-trajectories} gives the individual trajectories
summarized in \Cref{fig:decision-making}.

Across the 30 plotted trajectories, runs achieving at least twice the
strongest reference value reach a usable
end-to-end replay in a median of 0.90 hours, compared with 2.05 hours for
runs between one and two times the reference and 2.96 hours for verified
runs below it (\Cref{fig:first-replay}). A usable replay executes the
workload and yields positive latency-qualified value. Earlier measurement
leaves more of the development budget for subsequent design and testing.

\subsection{Verification Coverage and Continued Applicability}
\label{app:verification-coverage}
Verification annotation distinguishes long-context tests, concurrent
requests, state reuse, natural stopping, stopping under
load or extended contexts, and cold restarts. Coverage records executed
checks and their applicability after subsequent edits. The classification
distinguishes checks on the submitted tree, earlier checks followed by
relevant edits, and absent coverage.

The failure reconstruction links observed symptoms, agent responses,
repairs, and later regressions, including symptoms dismissed or left
unexamined and faults recurring after repair.

Post-run review applies the task's prescribed precision and inference
semantics to all submitted systems.

\Cref{fig:verification} records natural-stopping checks under load
or long context for 5 of 33 entries, and prior related symptoms for
8 of the 11 not-passed entries.

\subsection{Controlled Design and Load Comparisons}
\label{app:controlled-design}
\label{app:capacity-comparisons}
\label{app:design-analysis}
Frozen systems are replayed on three common seeds selected before the
comparison. Primary matched-system comparisons include Opus 5.5 and Opus 5
on Qwen, Sonnet 5.5 and MiMo on Gemma, and each deployment's SGLang
reference. Additional comparisons include Gemma Opus 5.5, Opus 5, and
Qwen3.8-Max-0902. Replays retain the task's workload, targets, and accounting.
Repeated replays quantify workload variation for each frozen system.

\Cref{tab:version-comparisons} reports fixed-load version comparisons
with qualified-value gains across their matched seeds. Each pair measures
the complete set of implementation changes between those versions.

\begin{table}[t]
\centering
\fontsize{9}{11}\selectfont
\setlength{\tabcolsep}{4pt}
\renewcommand{\arraystretch}{1.12}
\begin{tabular*}{\columnwidth}{@{\extracolsep{\fill}}ll@{}}
\toprule
Versions & Matched-seed value ratios \\
\midrule
\multicolumn{2}{l}{Opus 5.5 / Qwen 9B, $c=16$} \\
v1 $\to$ v2 & 1.88, 1.32 \\
\midrule
\multicolumn{2}{l}{Opus 5 / Qwen 9B, $c=2$} \\
v3 $\to$ v4 & 1.90, 1.67, 1.44 \\
\midrule
\multicolumn{2}{l}{Sonnet 5.5 / Gemma E4B, $c=32$} \\
v3 $\to$ v4 & 1.24, 1.02, 1.05 \\
\bottomrule
\end{tabular*}
\caption{Selected qualified-value gains from successive versions at fixed concurrency.
Each value corresponds to a scoreable matched-seed pair.}
\label{tab:version-comparisons}
\end{table}

Load comparisons vary concurrency while retaining the implementation and
common seeds. Sonnet and MiMo on Gemma use 2, 10, and 32
sessions; Sonnet's mean qualified value increases from 1.30 to 2.97.
Additional curves
compare Gemma Opus 5 at 8, 24, and 40 sessions, Gemma
Opus 5.5 at 10 and 40, and the two Qwen Opus systems at one, four,
and 16. Warmup completion is recorded separately from scored replay.
\Cref{fig:value-analysis} plots these load comparisons alongside
final-evaluation measurements.

\subsection{Observation Windows and Workload Sampling}
\label{app:controlled-measurement}

\begin{figure}[t]
\centering
\includegraphics[width=\textwidth]{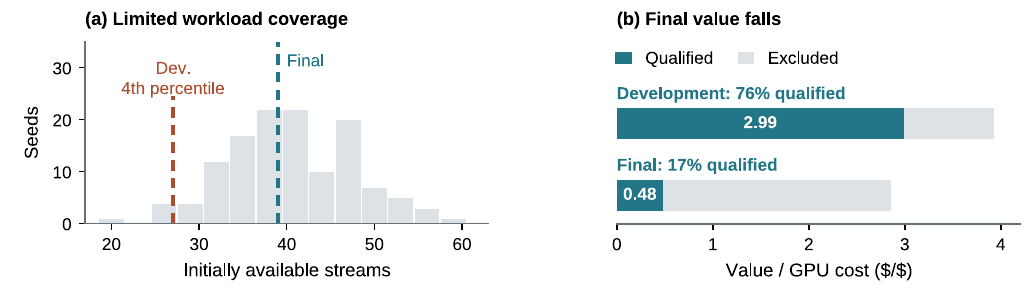}
\caption{Workload coverage and latency-qualified value in an Opus 5.5
27B development trajectory. (a) Initially available streams across 128
additional workload seeds, with development and final draws marked.
(b) Qualified and excluded token value for the same system at 16
sessions under the development and final replays.}
\label{fig:feedback-decisions}
\end{figure}
\Cref{fig:measurement-conditions} rescores prefixes of the same
replay, holding implementation, concurrency, and sampled sessions fixed.
It shows deployment-level mean window ratios in panel (a) and across-seed
coefficients of variation in panel (b).

We profile initial stream availability across 128 workload seeds.
For the 16-session Opus 5.5 27B run, the development draw lies near
the fourth percentile and the final draw near the median
(\Cref{fig:feedback-decisions}). Initial availability measures the
independent streams ready to advance when replay begins.

The development replay qualifies 76.0\% of token value and scores 2.99
dollars per dollar. Under the final draw, the qualified share falls to
16.8\% and the score to 0.48. Profiling the session distribution therefore
helps interpret the favorable development result and identify workload
conditions that warrant further testing.

For the current window curves, each seed's prefix score is normalized by
its 1,800-second score; system-level seed medians are then averaged within
each deployment. Panel (b) averages system-level across-seed coefficients
of variation. The 27B agent submissions each contribute one recorded
replay to panel (a); only its two multi-seed references contribute to
panel (b). Deployment summaries weight each system equally.

\section{Finding-Guided Comparisons}
\label{app:finding-guided}
We report serving improvements and verification checks from interventions
suggested by the three findings on agent-built servers.
\Cref{tab:finding-guided} lists the serving improvements. Each pair
uses fresh common replay seeds, the same deployment, and the 1,800-second
final replay followed by independent correctness checks in both arms.
All reported serving arms pass verification. Ratios compare qualified
value $D$, gross value $G$, and qualified share $f=D/G$.
Verification repairs are assessed by targeted correctness tests.

\begin{table*}[t]
\centering
\fontsize{9}{11}\selectfont
\setlength{\tabcolsep}{4pt}
\begin{tabular*}{\textwidth}{@{\extracolsep{\fill}}llrrrrr@{}}
\toprule
Comparison & Seed & $D$ before & $D$ after & $D$ ratio & $G$ ratio & $f$ ratio \\
\midrule
Capacity & D & 3.533 & 4.761 & 1.348 & 1.110 & 1.214 \\
 & E & 4.294 & 4.960 & 1.155 & 1.074 & 1.076 \\
\midrule
Draft length & D & 2.240 & 2.308 & 1.030 & 1.030 & 1.000 \\
 & E & 2.450 & 2.544 & 1.039 & 1.029 & 1.009 \\
\midrule
Replay feedback & F & 0.786 & 0.888 & 1.129 & 1.112 & 1.015 \\
 & G & 0.878 & 1.148 & 1.308 & 1.243 & 1.052 \\
\bottomrule
\end{tabular*}
\caption{Selected serving improvements from finding-guided changes on fresh common seeds.
Capacity compares Qwen v1 to v6 at $c=16$; draft length compares Gemma
$K=3$ to $K=7$ at $c=40$; replay feedback compares two resumed Opus 5
Gemma servers. Ratios are after/before; all arms pass verification.}
\label{tab:finding-guided}
\end{table*}

\subsection{Capacity at Fixed Load}
Opus 5.5's Qwen3.5 9B v1 and v6 run on $2\times$L40 at $c=16$.
The v1$\to$v6 bundle improves qualified value on both confirmation
seeds, by $1.35\times$ and $1.15\times$. Gross value rises by
$1.11\times$ and $1.07\times$, while qualified share rises by
$1.21\times$ and $1.08\times$: the gain combines greater throughput and
fewer latency losses. The v1$\to$v6 redesign extends the earlier
fixed-load v1$\to$v2 comparison in \Cref{tab:version-comparisons}.

\subsection{Workload Feedback and Design Decisions}
The corrected prompt-lookup simulator matches prompt and continuation
lengths in a crossed comparison of real and synthetic text over 131
requests. At draft length $K=7$, pooled emitted tokens per step are
2.635 for real prompts and replies, 1.744 for synthetic prompts with real
replies, 1.246 for real prompts with synthetic replies, and 1.244 for
fully synthetic text. Emitted tokens include accepted draft tokens plus
one verification token. With an assumed incremental draft-cost parameter
$\alpha=0.01$, workload replies select $K=7$, whereas fully synthetic
text selects $K=3$. The pooled draft-length selection holds over
$\alpha\approx0.004$--$0.047$.

Serving tests change only $K$ on Opus 5.5's Gemma v7 at $c=40$.
The workload-selected $K=7$ increases qualified value by 3.0\% and 3.9\%
on the two fresh seeds, mainly through gross value.

To test feedback's effect on agent behavior, we resume Opus 5's Gemma
v2 checkpoint. Two continuations use the same model, task, checkpoint,
and two-hour budget, with development-only workload data. The feedback
arm receives its own 205 development-replay replies and their prompt-lookup
measurements, then re-adopts speculation. Its
submitted server raises qualified value by $1.13\times$ and
$1.31\times$ on fresh common seeds F and G.

\subsection{Verification and Repair}
Opus 5.5's Gemma cache repair masks invalid cache loads, initializes cache
pools, and uses NaN-safe argmax. The full repair passes the 420-page
concurrent check (1/1), all three 330-page checks exercising preemption
and recovery (3/3), and the 330-page check with preemption disabled (1/1).
The submitted v7 also passes all three targeted checks (3/3).
Masking invalid cache loads and zeroing cache pools each passes its
targeted attempt; the full repair passes all three repetitions.
These checks verify correct outputs while exercising the server's cache
management and recovery paths.

\end{document}